\documentclass{article}
\usepackage{ijcai21}

\usepackage{times}
\usepackage{soul}
\usepackage[hidelinks]{hyperref}
\usepackage[utf8]{inputenc}
\usepackage[small]{caption}
\usepackage{amssymb,amsfonts}
\usepackage[ruled, vlined, linesnumbered]{algorithm2e}
\usepackage{amsmath,accents}
\usepackage[noend]{algpseudocode}
\usepackage{graphicx}
\usepackage{textcomp}
\usepackage{xcolor}
\usepackage{color}
\usepackage{tabularx,multirow}
\usepackage[inline]{enumitem}
\usepackage[capitalise]{cleveref}
\usepackage{tcolorbox}
\usepackage{empheq}
\usepackage{multirow}
\usepackage{subcaption}
\usepackage{xspace}
\usepackage{booktabs}
\usepackage{bm}
\usepackage{dblfloatfix}
\usepackage{xurl}
\usepackage{amsthm,lipsum}
\theoremstyle{definition}
\newtheorem{definition}{Definition}
\theoremstyle{remark}

\usepackage{color, colortbl}

\let\oldnl\nl
\newcommand{\nonl}{\renewcommand{\nl}{\let\nl\oldnl}}

\tcbset{arc=0mm,size=fbox,attach title to upper={\ ---\ },coltitle=black}

\usepackage{array}
\newcolumntype{L}[1]{>{\raggedright\let\newline\\\arraybackslash\hspace{0pt}}m{#1}}
\newcolumntype{C}[1]{>{\centering\let\newline\\\arraybackslash\hspace{0pt}}m{#1}}
\newcolumntype{R}[1]{>{\raggedleft\let\newline\\\arraybackslash\hspace{0pt}}m{#1}}

\crefformat{section}{\S#2#1#3} 
\crefformat{subsection}{\S#2#1#3}
\crefformat{subsubsection}{\S#2#1#3}

\newcommand{\tra}{TRA\xspace}

\title{Loss Tolerant Federated Learning}

\begin{document}

\author{
Pengyuan Zhou$^1$\and
Pei Fang$^2$\And
Pan Hui$^{1,3}$\\
\affiliations
$^1$University of Helsinki\\
$^2$Tongji University\\
$^3$Hong Kong University of Science and Technology\\
\emails
pengyuan.zhou@helsinki.fi,
greilfang@gmail.com,
pan.hui@helsinki.fi
}
\maketitle

\begin{abstract}
Federated learning has attracted attention in recent years for collaboratively training data on distributed devices with privacy-preservation. The limited network capacity of mobile and IoT devices has been seen as one of the major challenges for cross-device federated learning. Recent solutions have been focusing on threshold-based client selection schemes to guarantee the communication efficiency. However, we find this approach can cause biased client selection and results in deteriorated performance. Moreover, we find that the challenge of network limit may be overstated in some cases and the packet loss is not always harmful.

In this paper, we explore the loss tolerant federated learning (LT-FL) in terms of aggregation, fairness, and personalization. We use ThrowRightAway (TRA) to accelerate the data uploading for low-bandwidth-devices by intentionally ignoring some packet losses. The results suggest that, with proper integration, TRA and other algorithms can together guarantee the personalization and fairness performance in the face of packet loss below a certain fraction (10\%–30\%).
\end{abstract}


\section{Introduction}\label{sec:introduction}
With the popularization of the mobile and wearable devices, smart activity learning applications have been prominently used by consumers and in turn generate more user data.
Despite the potential to act as effective data sources for machine learning tasks, the training of machine learning models for mobile and wearable applications usually demands data far more than each individual device collects. Currently, aggregating user data in the cloud for big data analysis is the de facto solution. However, privacy concerns have spawned a series of policies that limit data collection and storage only to consumer-consented and absolutely necessary usage~\cite{lim2020federated}. For example, most data collected from mobiles and wearables are subject to data protection regulations such as European Commission’s General Data Protection Regulation (GDPR)~\cite{custers2019eu} and Consumer Privacy Act (CCPA) in USA~\cite{ccpa}. Such regulations make it harder to aggregate user data for large scale data analysis.

In face of the above challenge, federated learning rises as a new distributed paradigm where multiple clients collaboratively train a model without revealing private data. Based on whether the clients are different organization or a large number of mobile IoT devices, federated learning is divided into cross-silo and cross-device,  
Mobile and wearable devices as the major participants, cross-device federated learning faces challenges from stateless and unreliable clients. Moreover, communication seems to be another bottleneck as the operations of cross-device federated learning systems largely rely on Wi-fi~\cite{bonawitz2019towards} or slower communication networks. 

With the concern on improving the communication efficiency, most of the recent works propose or assume a threshold to select clients with sufficient network capacities. 
However, such proposals inevitably cause data shifts during client selection. Although very recently researchers have proposed fairness schemes specifically for federated learning aggregation, data shift occurring at the beginning of client selection has been overlooked.
Consequently, the performance of federated learning is impacted.

In this paper, we reexamine the network limit challenge and the threshold-based approach to answer the following questions:
\begin{enumerate*}
    \item Is the challenge overstated?
    \item What is the drawback of threshold-based client selection approach?
    \item Are there better alternative solutions?
\end{enumerate*}
Concretely, we make the following contributions in this work:
\begin{enumerate}[wide = 0pt]
\item \textit{Bottlenecks.} We conduct a trace-driven analysis and learn that the network limit challenge may be overstated in some aspects. Meanwhile, we identify an overlooked bias potentially caused by threshold-based client selection. We further analyze its impact on the performances of the state-of-the-art algorithms in the fields of aggregation, fairness and personalization~(\cref{sec:problem}).
\item \textit{Loss tolerance.} We explore the loss tolerant federated learning (LT-FL) by using ThrowRightAway (TRA) to ignore some of the lost packets on purpose. As its name indicates, TRA ``throws'' away lost packets to accelerate the uploading of low-bandwidth-devices thus enable fully fair selection to all the clients, while avoiding straggling impact caused by retransmissions. For specific algorithms, TRA tweaks the aggregation algorithm to compensate for the lost information~(\cref{sec:system}).
\item \textit{Applicability.} We apply TRA to aggregation, personalization, and fairness, by respectively integrating it with the state-of-the-art algorithms in each subfield. The empirical evaluation results show that TRA improves the performance of the paired algorithms in threshold-based client selection settings. We implement the combinatorial algorithms based on the state-of-the-art codebases and open source here~\footnote{\url{https://github.com/Greilfang/Loss-Tolerant-Federated-Learning}}.
\end{enumerate}

\section{Background and Motivation}\label{sec:background}
In this section, we describe the background and drawback of threshold-based client selection scheme and state the motivations of this work.

\subsection{Fair Client Selection} \label{ssec:motivation-fairness}

\begin{figure}[!t]
    \centering
    \includegraphics[width=.7\linewidth]{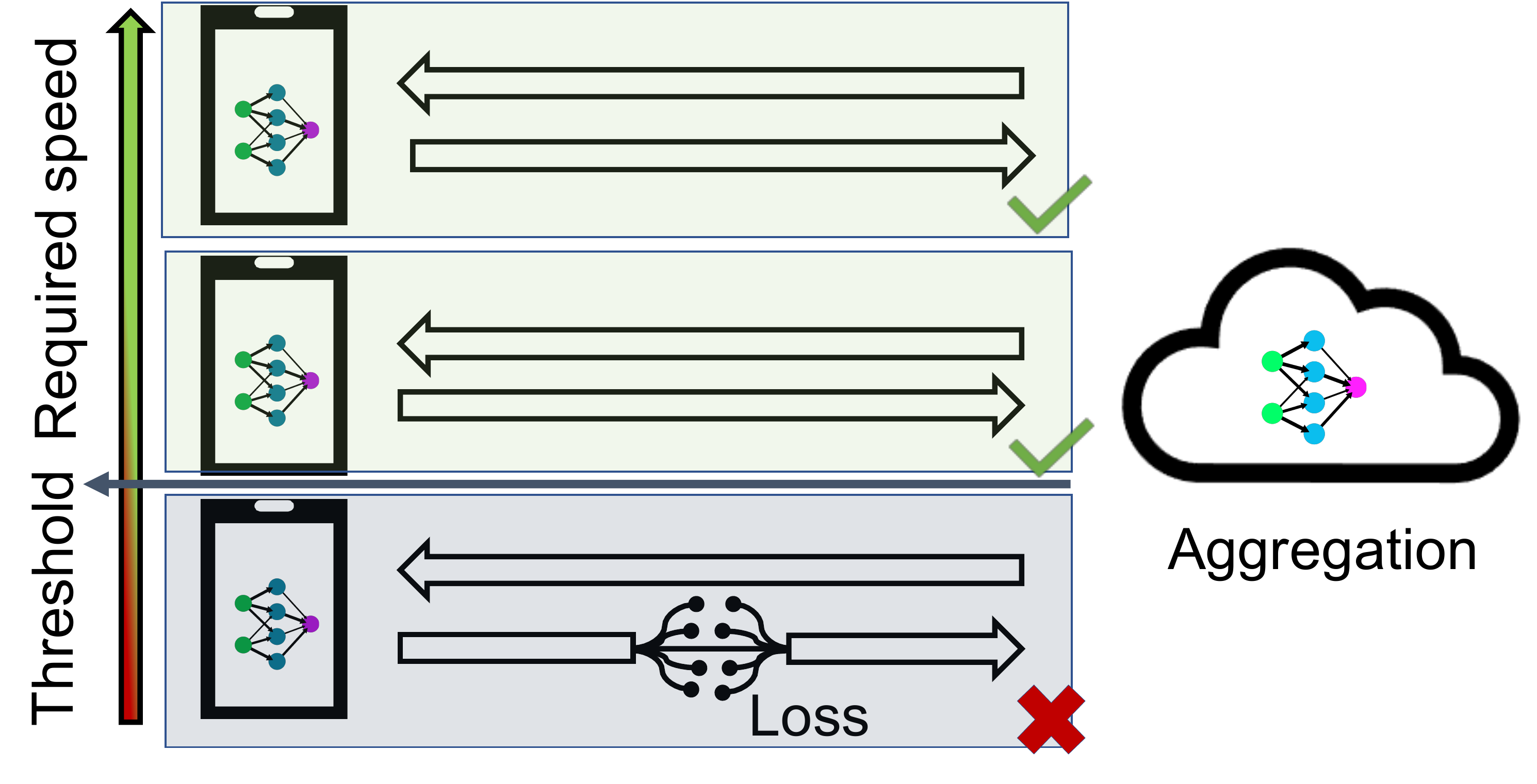}
    \caption{Threshold-based schemes select clients with better network conditions, e.g., higher speed/bandwidth, to avoid packet loss and stragglers during aggregation.}
    \label{fig:losscase}
\end{figure}

As noted by Bonawitz et al.~\cite{bonawitz2019towards}, the FedAvg~\cite{mcmahan2017communication} model aggregation protocol's assumption about equitable participation of all devices is not the case in practice. Consequently, fairness~\cite{barocas2017fairness,fang20achieving,lyu2020collaborative} is impacted and results in bias. For instance, to avoid packet error and client drop~(\cref{fig:losscase}), cross-device federated learning systems commonly use transmission speed and battery status as criteria for mobile client selection.
In such cases, the clients with more packet errors and drops are unlikely be taken into model aggregation. Even worse, \textit{users consistently having worse networking conditions may never be represented in training which leads to a biased model.} The aggregation approaches with only model weights taken into account have been proved unable to tackle such challenges~\cite{karimireddy2020scaffold,lin2020federated}.

Here we summarize common factors for bias in federated learning as: (1) \textit{over-represented}, (2) \textit{under-represented}, (3) \textbf{\textit{never-represented}}.
(1) and (2) have been partly solved with approaches targeting training procedure bias such as AFL~\cite{mohri2019agnostic} and q-FedAvg~\cite{li2019fair}. AFL minimizes the maximum loss incurred on the worst-performing devices as a classical minimax problem. q-FedAvg generalizes AFL by allowing for a flexible trade-off between fairness and accuracy. Although these approaches promote accuracy equity among participaed devices through the mitigation of training procedure bias, still they can not solve bias caused by unfair client selection in (3), as also noted by the authors of AFL. 
\subsection{Inspirations}
There have been recent works focusing on different techniques suggesting  intentionally ``losing'' some information to avoid latency for better communication efficiency, and ease the gap between demanded and actual network capacities. For instance, some related works have proposed to use lossy compression to reduce the transferred data volume. The authors in~\cite{konevcny2016federated,dong2020cdc} perform lossy compression on the model updates using both structured and sketched updates. The main idea is to learn from a restricted space or upload a compressed model. Authors in~\cite{caldas2018expanding} focus on the server-to-client communication and similarly applies a lossy compression scheme with less frequent updates. The authors in~\cite{xia2019rethinking} tapped into the loss tolerance potential in distributed machine learning, which show its bounded loss tolerance of via evaluations. These works inspire us to explore the network data loss tolerance of cross-device federated learning. The differences between our work and aforementioned works are twofold: \begin{enumerate*}
    \item We propose a loss-tolerant scheme not only to address communication efficiency, but also to guarantee fairness during client selection.
    \item We look deeper into the potential of a loss-tolerant scheme by integrating it into two state-of-the-art fairness algorithms regarding fairness and personalization respectively and show performance improvements in different aspects.
\end{enumerate*}
\begin{tcolorbox}
\textit{Note:} The network threshold for selection can be bandwidth, transmission speed, or packet loss, or their hybrids. In this work we use \tra to convey different network constraints to general packet loss. Specifically, \tra intentionally drops lost packets in avoidance of re-transmission to allow a client with slower network to upload local models within a jointly-decided period with other clients.
\end{tcolorbox}

\section{Problem Study}\label{sec:problem}
In this section, we analyze the problems mentioned in~\cref{sec:background} in detail. First we learn the disparate networking conditions by analyzing a real-world dataset and discover its biased impact on client selection. Then we show how the state-of-the-art approaches regarding fairness and personalization for federated learning suffer from the data shift due to the biased selection.

\subsection{Mobile Network Conditions}\label{ssec:bottleneck-condition}

\begin{figure}[!t]
    \centering
    \includegraphics[width=\linewidth]{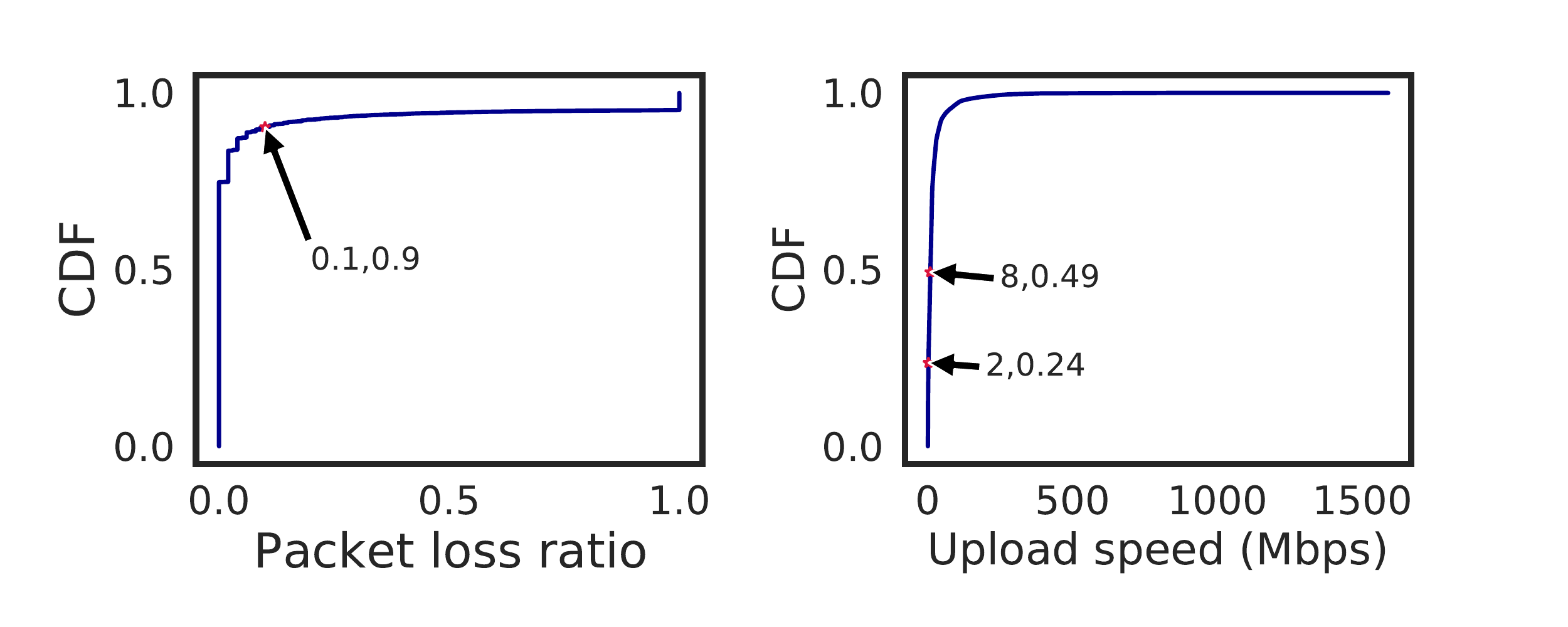}
    \caption{Network conditions analysis.}
    \label{fig:problem-network}
\end{figure}

We use a mobile broadband dataset provide by FCC~\cite{fccfix} to study the mobile network conditions of users. We select data from the “Download speed and upload speed” category in 2019 Q1 \& Q2 collection. The data is measured via Android and iOS applications installed in the user phones. It contains uploading traces from thousands of volunteered participants, recording the average received packets, lost packets and throughput. Note that the throughput was collected as the sum of speeds during saturated streams, hence it can represent the max speed. We calculate the packet loss by dividing the lost packets by the sum of received and lost packets.  After processing the trace according to unique identifiers, the cumulative distributions of the average packet loss ratio and upload speed are shown in \cref{fig:problem-network}. It shows that 90\% of the users have packet loss ratio $< 0.1$ and 76\% of the users have upload speed $> 2$ Mbps. Therefore the majority of the users have sufficient network capacities required by common federated learning systems. However, the upload speeds vary tremendously across users. For instance, 24\% of the users have upload speed $< 2$ Mbps while 51\% of the users have upload speed $>8$ Mbps.  
Transmission speed is an important metric during client selection and has been adopted by both industrial and academic works~\cite{nishio2019client,openmined}. For instance, Openmined sets 2 Mbps as the default upload speed threshold for client selection. As we found out from the data analysis, a considerable part of users may fail to meet the network threshold thus would be \textit{never-represented} in the model aggregation and thus being excluded by the system.
\begin{tcolorbox}
\textit{Takeaway}: The trace-driven analysis shows that the network conditions of most mobile clients are not so ``limited'' and ``challenging'' as most related works assumed. However, the tremendously varied upload speeds may indeed cause biased client selection in threshold-based settings.
\end{tcolorbox}
\subsection{Impacts}\label{ssec:impact}
Following the takeaway in~\cref{ssec:bottleneck-condition}, we investigate the impact of biased selection led by threshold-based settings. We define the essential terms in our investigated problem as follows.
\begin{definition}[\textit{\textbf{Eligible client}}]\label{definition-client}
An eligible client is one that meets the required network threshold to participate in federated learning aggregation.
\end{definition}
\begin{definition}[\textit{\textbf{Eligible ratio}}]\label{definition-ratio}
Eligible ratio is the proportion of the eligible clients out of all the clients.
\end{definition}
In threshold-based settings, only the eligible clients within the eligible ratio may be selected for aggregation. As some users have lower network capacities than the threshold~(\cref{fig:problem-network}), the system only can choose eligible clients for aggregation and generate bias and result in models with discrimination.
For the completeness of the work, we adjust the eligible ratios between 100\%, 90\%, 80\%, and 70\% in the evaluation of the paper. More specifically, we investigate the impacts on aggregation, fairness, and personalization, respectively. In the rest of the evaluation, we use the synthetic datasets generated following the process described in the experiment detail of q-FedAvg, where $\alpha$ and $\beta$ allow the precise manipulation of the degree of heterogeneity. Increasing the values of $\alpha$ and $\beta$ result in higher statistical heterogeneity. We use the same datasets for both bottleneck analysis and evaluation for consistency.

\begin{figure}[!t]
    \centering
    \includegraphics[width=.5\linewidth]{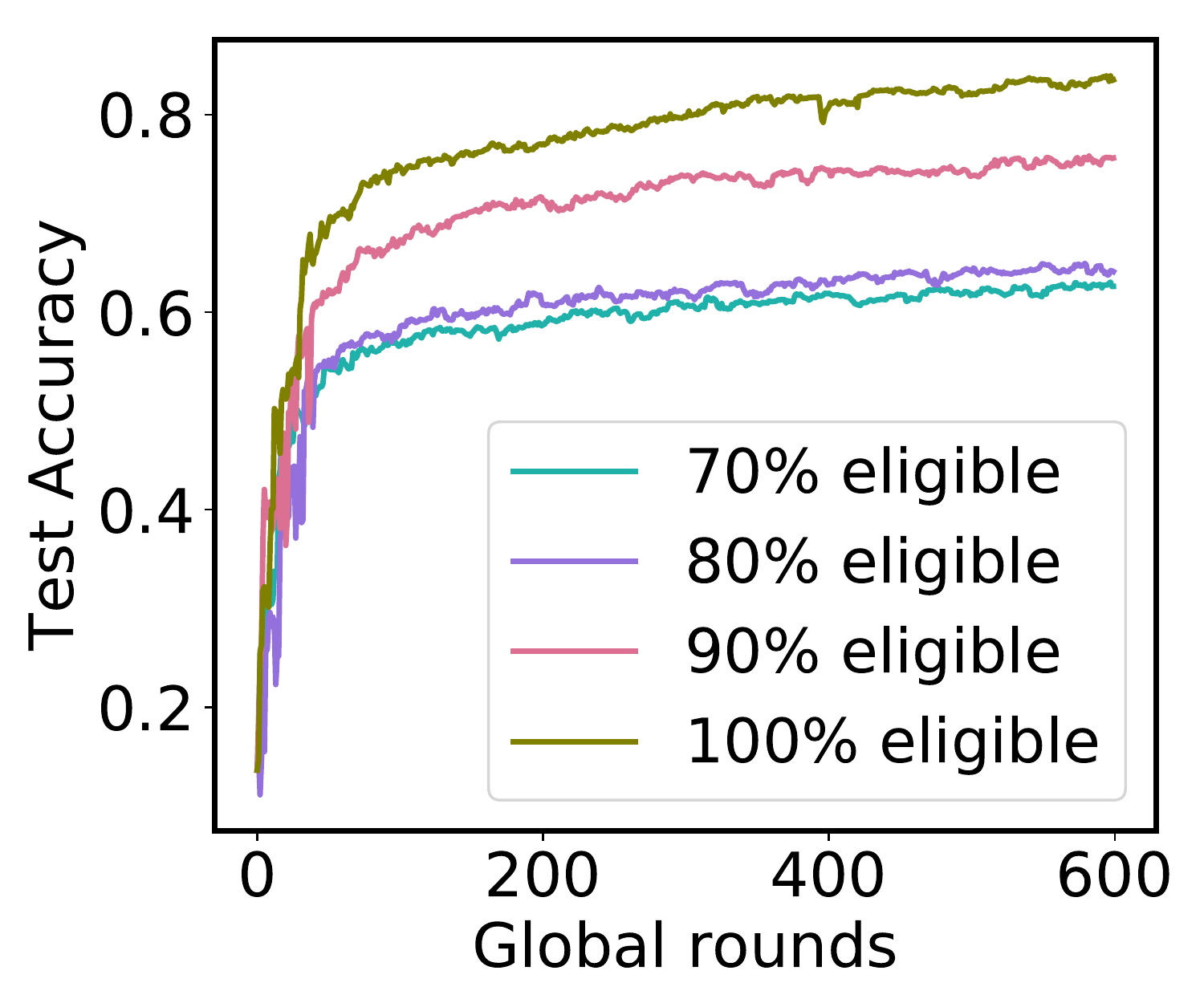}
    \caption{Impact of biased selection on aggregation (FedAvg). The dataset is Synthetic(0.5,0.5).}
    \label{fig:bottleneck-aggregation}
\end{figure}

\paragraph{Aggregation}
First we examine the impact of biased selection on aggregation. We target at the prevailing and common FedAvg, which evenly averages the selected clients' models.  As~\cref{fig:bottleneck-aggregation} shows, smaller eligible ratios have higher impacts on the model performance. The final model accuracy of FedAvg with eligible ratios of 100\%, 90\%, 80\%, and 70\%, are 83.52\%, 75.60\%, 64.10\%, and 62.60\%. For the users in~\cref{fig:problem-network}, the model accuracy would decrease around 10\% if using 2 Mbps as the selection threshold.

\begin{table}[!t]
\caption{Impact of biased client selection on fairness (q-FedAvg). Threshold (TH) indicates whether considering the 70\% eligible ratio. Best/Worst 10\% indicate the top 10\% best/worst accuracies. }
\label{tab:bottleneck-fairness}
\scalebox{0.9}{
\begin{tabular}{cllll}
\specialrule{1.3pt}{1pt}{1pt}
\multicolumn{1}{c|}{\textbf{Dataset}}                                                                    & \multicolumn{1}{l|}{\textbf{TH}} & \textbf{Average} & \textbf{Best/Worst 10\%} & \multicolumn{1}{l}{\textbf{Variance}} \\ \hline
\multicolumn{1}{c|}{\multirow{2}{*}{\begin{tabular}[c]{@{}c@{}}\textbf{Synthetic}\\ \textbf{(i.i.d)}\end{tabular}}} & \multicolumn{1}{l|}{} & 72.47\%     & 91.85\% / 43.19\%              & \multicolumn{1}{l}{179}        \\ \cline{2-5} 
\multicolumn{1}{c|}{}                                                                           & \multicolumn{1}{l|}{\checkmark}  & 68.67\%     & 94.25\% / 36.30\%              & \multicolumn{1}{l}{245}        \\ \hline
\multicolumn{1}{c|}{\multirow{2}{*}{\begin{tabular}[c]{@{}c@{}}\textbf{Synthetic}\\ (\textbf{0.5,0.5})\end{tabular}}} & \multicolumn{1}{l|}{ } & 66.21\%     & 98.30\% / 22.51\%              & \multicolumn{1}{l}{536}        \\ \cline{2-5} 
\multicolumn{1}{c|}{}                                                                           & \multicolumn{1}{l|}{\checkmark}  & 52.81\%     & 99.79\% / 0  & \multicolumn{1}{l}{1350}        \\ \hline
\multicolumn{1}{c|}{\multirow{2}{*}{\begin{tabular}[c]{@{}c@{}}\textbf{Synthetic}\\ (\textbf{1,1)}\end{tabular}}} & \multicolumn{1}{l|}{ } & 64.17\%     & 100\% / 7.67\%               & \multicolumn{1}{l}{937}        \\ \cline{2-5} 
\multicolumn{1}{c|}{}                                                                           & \multicolumn{1}{l|}{\checkmark}  & 55.24\%     & 100\% / 0               & \multicolumn{1}{l}{1439}        \\ \hline
\end{tabular}
}
\end{table}

\paragraph{Fairness}
As noted in~\cref{ssec:motivation-fairness}, existing schemes improve fairness for \textit{over-represented} and \textit{under-represented} clients, but fail to serve the \textit{never-represented} clients. To validate this argument, we reproduce the evaluations of q-FedAvg using the code and default hyperparameters provided by the authors of q-FedAvg.  
We use a 70\% eligible ratio to get the bottleneck performance. 
We adjust the distribution of training sample data on each device (from i.i.d data to non-i.i.d data) to comprehensively test the degradation of both accuracy and fairness performance caused by biased client selection.
\cref{tab:bottleneck-fairness} shows that the performances of q-FedAvg are impacted due to biased selection with both i.i.d and non-i.i.d data distributions. Non.i.i.d data presents larger performance degradation than i.i.d data in terms of both accuracy and fairness. 

\paragraph{Personalization}
Existing approaches either trai n a new deep neural network (transfer learning)~\cite{9076082}, with loss function measuring the heterogeneity for local and global models, other than the one for the task. In resource-intensive cases, transfer learning reduces the model size so that a device can simultaneously hold two transferable models, but its advantage over a single larger model requires further explorations. 
Per-FedAvg~\cite{fallah2020personalized} looks for an initial shared model that clients can easily adapt via a few gradient descents with respect to their own data.
pFedMe~\cite{dinh2020personalized} adds constraints into the loss function of global training and shows outperformance of Per-FedAvg.
Therefore, we use pFedMe as the target to examine the impact of biased selection on personalization performance.

\begin{figure}[!t]
    \centering
    \includegraphics[width=\linewidth]{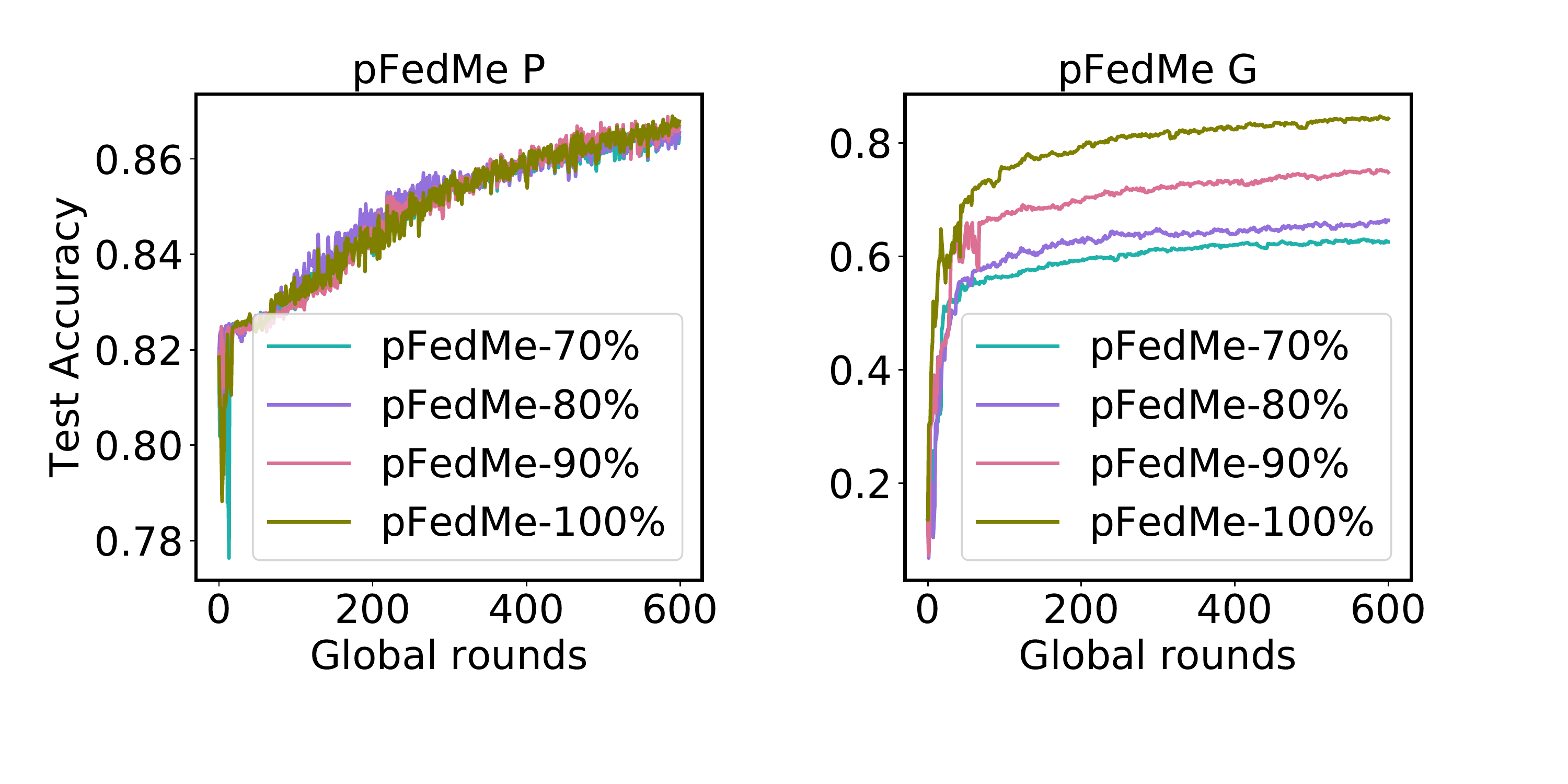}
    \caption{The impact of biased client selection on  personalized and global performance of pFedMe. Label \textbf{p} refers to the average local accuracy after personalization while \textbf{G} refers to the global accuracy. The dataset is Synthetic(0.5,0.5). We use the fine-tuned hyperparameters of Table. 1 in the paper of pFedMe.}
    \label{fig:bottleneck-personalization}
\end{figure}

\begin{figure}[!t]
    \centering
    \includegraphics[width=.5\linewidth]{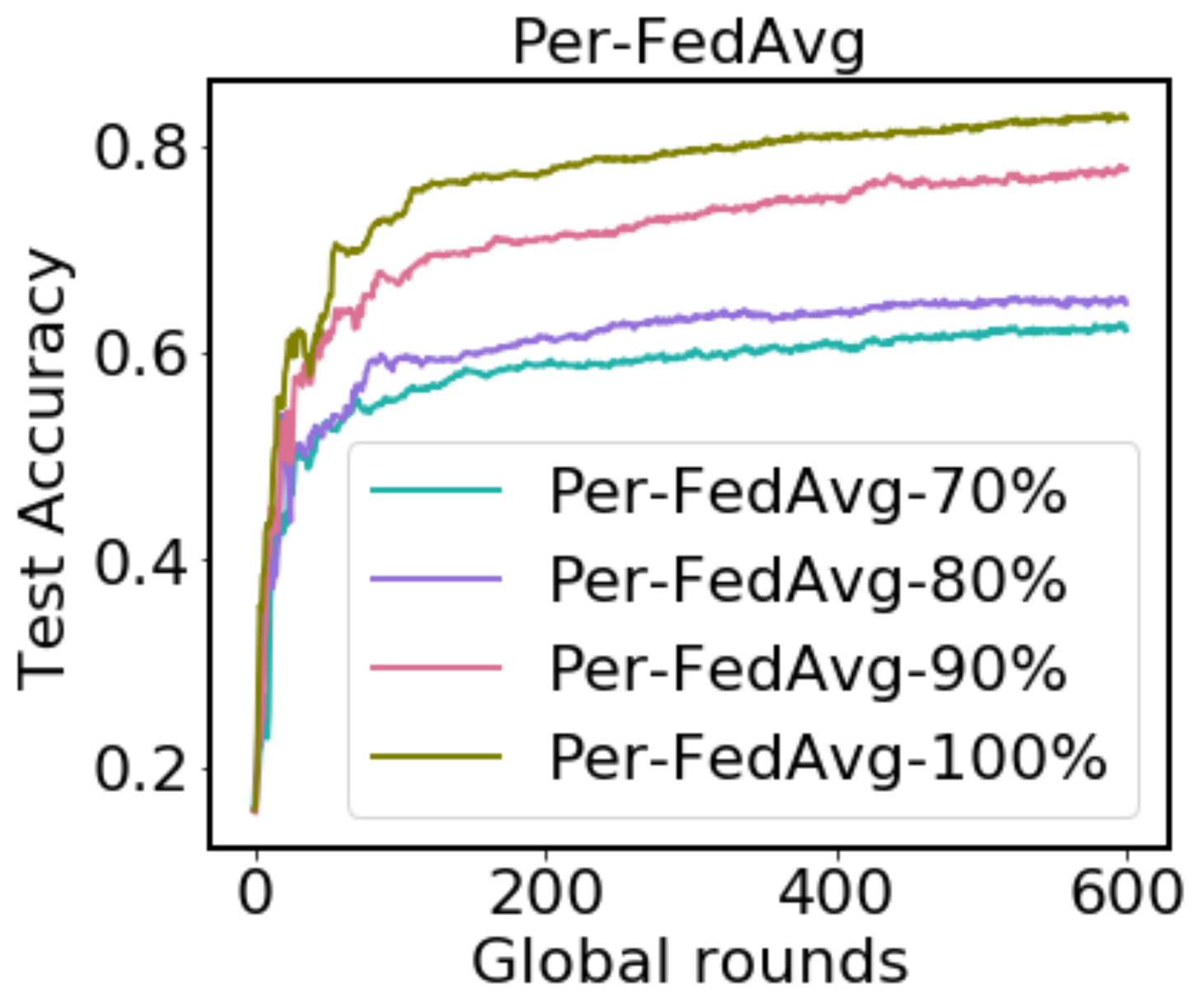}
    \caption{The impact of biased client selection on  Per-Fedavg. The dataset is Synthetic(0.5,0.5). We use the hyperparameters of Table. 1 in the paper of pFedMe.}
    \label{fig:bottleneck-personalization2}
\end{figure}

As shown in \cref{fig:bottleneck-personalization}, pFedMe shows resilient performance in its personalized model. However, the performance of the global model presents considerable degradation in lower eligible ratios.
We note that pFedMe achieves robustness on personalized model performance via more computation and power cost. Unlike most approaches selecting clients before local training, pFedMe let all clients do local training and then select some to upload. As such, its performance of personalized model is less depending on the convergence of the global model, while costing more computation and power of the client devices as a trade-off. For example, applying an eligible ratio to Per-FedAvg gets degraded performance as shown in~\cref{fig:bottleneck-personalization2}.

\begin{tcolorbox}
\textit{Takeaway}: The biased client selection caused by threshold-based selection can severely deteriorate the performance of aggregation, personalization, and fairness in the context of federated learning. Therefore, an alternative loss-tolerant selection scheme allowing fair participation is demanded.
\end{tcolorbox}
\section{ThrowRightAway}\label{sec:system}
In this section, we propose an alternative scheme to threshold-based schemes, to tackle the performance degradation caused by biased selection. 
\begin{figure}[!t]
    \centering
    \includegraphics[width=.7\linewidth]{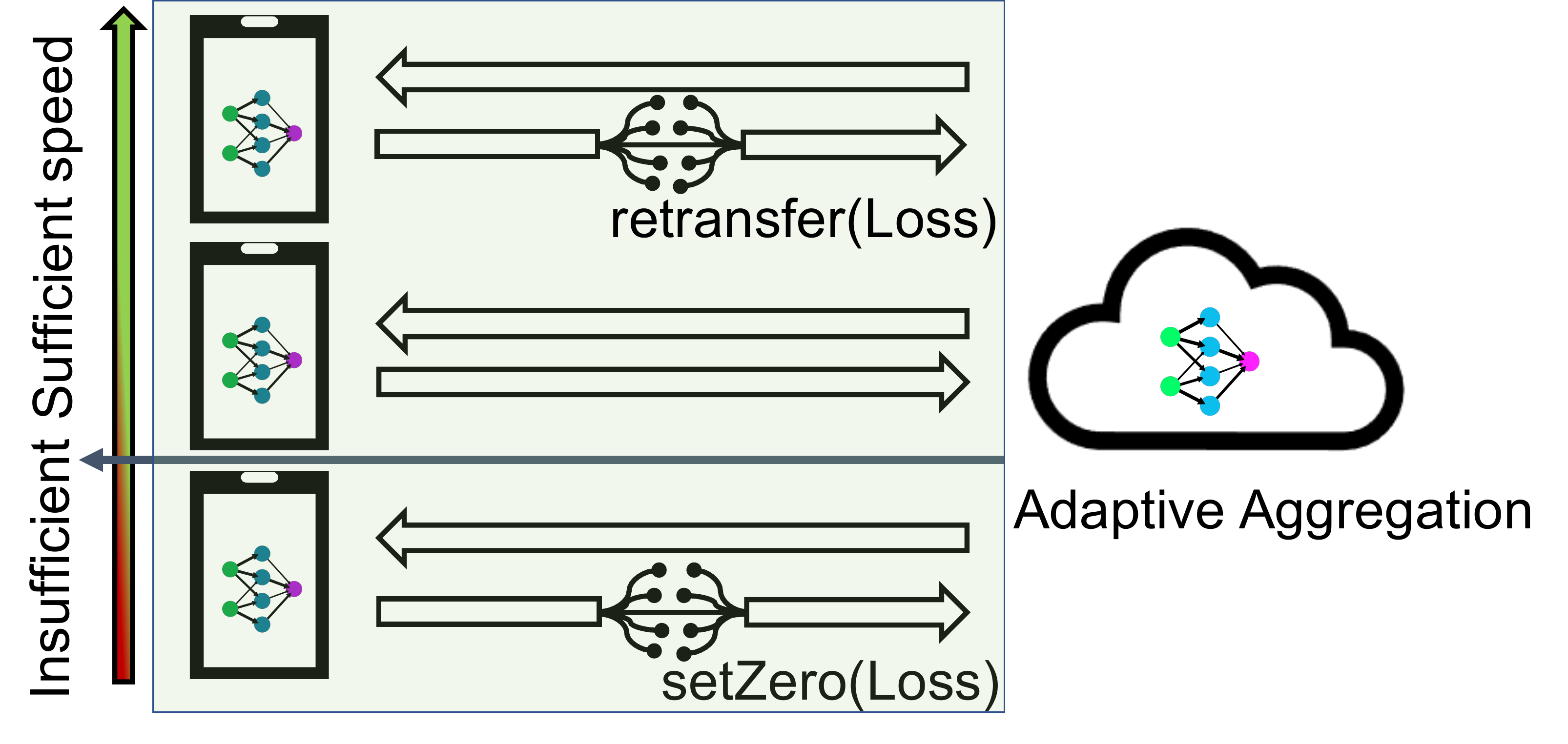}
    \caption{TRA allows clients participate in the aggregation regardless of network conditions..}
    \label{fig:tra}
\end{figure}

The authors in \cite{xia2019rethinking} have recently demonstrated that in contrary to common sense, data loss to an extent is not necessarily harmful in distributed learning systems. Through empirical evaluations, they discover that machine learning algorithms tolerate bounded data loss (10\%–35\% in their tests).
Inspired by the work, we propose to explore the loss tolerance in cross-device federated learning systems. We propose ThrowRightAway (\tra) scheme, i.e., the server accepts all clients as eligible participants even if some selected clients may have worse network capacities than requirement and undesired packet loss ratio during updates uploading~(\cref{fig:tra}). \tra is super lightweight and easy to implement. It can be integrated into different kinds of federated learning algorithms to augment their performances. \cref{alg:alg-tra} shows the skeleton of the integration of \tra and general federated learning algorithms.
\begin{figure*}[!t]
    \centering
    \includegraphics[width=.9\linewidth]{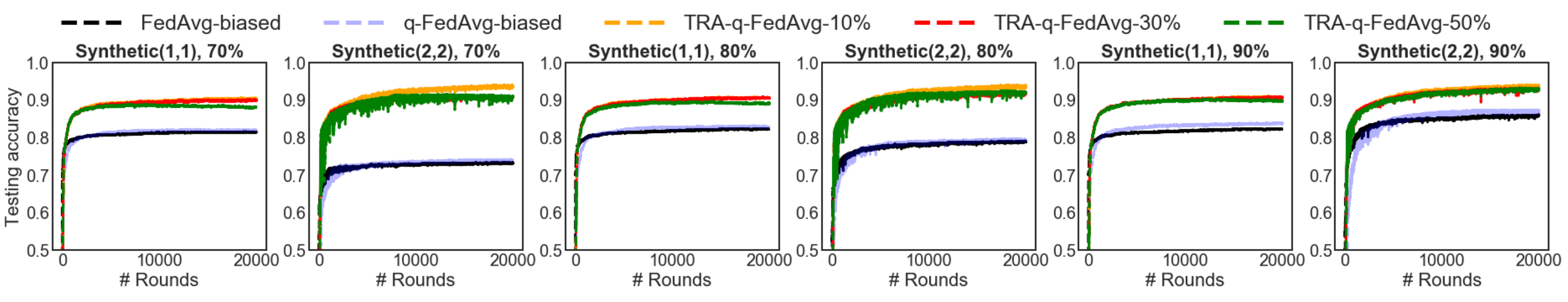}
    \caption{Sample based aggregations performance of biased FedAvg, q-FedAvg, and TRA-q-FedAvg on Synthetic(1,1) and Synthetic(2,2) datasets with 70\%, 80\%, and 90\% eligible ratios. TRA-a-FedAvg-X\% indicates the packet loss ratios (10\%, 30\%, 50\%).}
    \label{fig:evaluation-aggregation}
\end{figure*}

At the beginning of selection, each client compares its network condition with preset standards and sends a sufficiency investigation report to the server. The report contains only critical information, e.g., 0 or 1 to indicate insufficient or sufficient, thus adds negligible network load. After collecting the sufficiency reports of all willing-to-participate clients, the server classified the candidate clients into \textit{sufficient} and \textit{insufficient} based on the reports. Then the server randomly selects a number of clients regardless of the belonging groups and sends the global model. The clients send back updates after local training. Upon detecting loss, the server sends retransmission notification if the client belongs to the \textit{sufficient} group, or sets the lost data to zero directly otherwise. The rest of the processes follow the common federated learning flow.

We base the logic behind different loss operations on estimating whether retransmission would take long. For clients with sufficient network capacities, most likely the retransmission would only take very few times without affecting the aggregation pace. However for the ``insufficient'' clients, the retransmission is more likely to straggle and impact the system flow. Therefore instead of retransferring the lost packets to guarantee data integrity, \tra discards such delayed/lost packets, resets the lost data as 0, and records the data loss. After uploading finished, \tra uses the loss record to recalculate the sample space to achieve an adaptive aggregation. As such, \tra prevents the biased selection in threshold-based settings by safely ignoring some packet losses. The recalculation can be summarized as follows:
\begin{equation}\label{eq:recal}
    W_{agg} = \frac{1}{n}\sum_{i=1}^n W_i  + \frac{1}{m(1-r)}\sum_{i=1}^m \hat{W}_j
\end{equation}
$W_i$ and $\hat{W}_j$ are respectively model weights in $n$ users with sufficient and $m$ users with insufficient network capacities. $r$ indicates the package drop rate. Hence each weight $w$ in $\hat{W}$ has probability $r$ to be dropped and set zero. We denote $\mathbb{E}(W)=\mu$, then $\mathbb{E}(\hat{W}) = (1-r)\mathbb{E}(W) = (1-r)\mu$. When the clients' model have the same distributions, it is clear that $\mathbb{E}(W_{agg})=\mu$ and is equal to $\mathbb{E}(\frac{1}{n+m}\sum_{i=1}^{n+m} W_i)$, which is the expectation of the aggregated model without packet loss. Generally, $r$ ensures that, after the clients' model with packet loss being taken in, the aggregated model's weights would not be smaller than those without packet loss.

To validate the easy-to-integrate feature and performance of~\tra, we integrate~\tra with several state-of-the-art algorithms in different aspects of federated learning. \cref{sec:evaluation} shows the performance of the integrated algorithms.

\begin{algorithm}[!t]
\SetKwProg{proc}{thread}{:}{}
  \nonl \proc{Server}{
      collect(sufficiencyReport) \\
      categorize(sufficiencyGroup)\\
      Randomly selects a subset $S$ of the clients\\
      \For{t=1 to T-1}{
       \For{each client $k \in S$}{
            $\mathbf{x}_k^{t+1} \leftarrow$ \textit{Client}$(k,w^t)$\\
            \uIf{loss}{
                \uIf{sufficient}{retransfer(loss)}
                \Else{setzero(loss)}}}
      Update model $w^{t+1}$ via aggregation algorithm }
      Return($w^{T}$)
      }
  \nonl \proc{Client}{
         send(sufficiencyReport)\\
         $w_k^{t+1} \leftarrow E$ epochs of gradients $\leftarrow w^t$ \\
         Return($\mathbf{x}_k^{t+1}$) \Comment{$\mathbf{x}$ \textit{ is algorithm specific value}}
         }
\caption{The skeleton of integrating TRA into general federated learning algorithms.}
\label{alg:alg-tra}
\end{algorithm}

\section{Evaluation}\label{sec:evaluation}
To verify the performance of \tra, we redo the evaluations conducted in~\cref{ssec:impact}. 
We compare the performance of the algorithms limited by the threshold-based selection with the integrated algorithms. For realistic concern, we only consider nonconvex settings. Similarly with~\cref{ssec:impact}, we consider three eligible ratios, i.e., 70\%, 80\%, and 90\% which cause different degrees of biased client selection in threshold-based settings. For each eligible ratio, we consider a variety of packet loss ratios, i.e, 10\%, 30\%, and 50\%, for the \textit{insufficient} clients.

\begin{figure*}[!t]
    \centering
    \includegraphics[width=.9\linewidth]{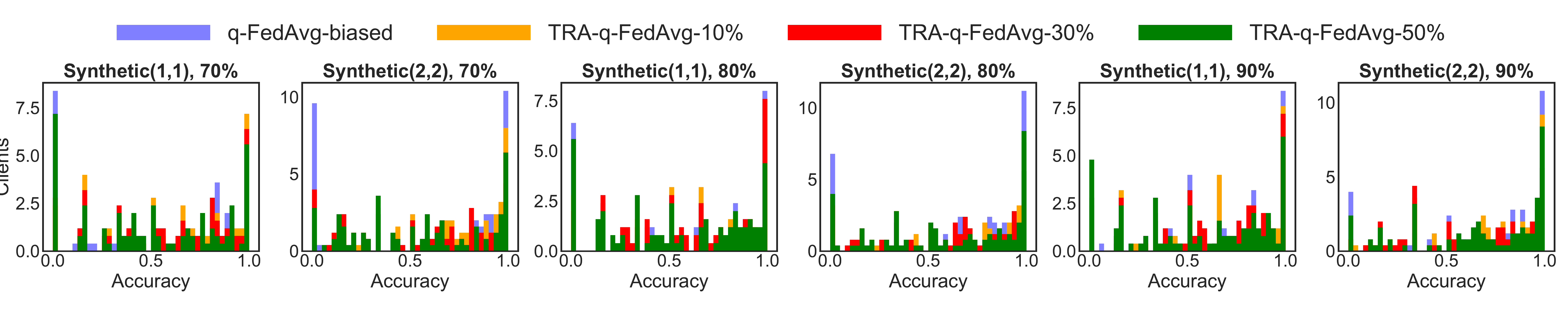}
    \caption{Fairness performance of biased q-FedAvg and TRA-q-FedAvg on Synthetic(1,1) and Synthetic(2,2) datasets with 70\%, 80\%, and 90\% eligible ratios. TRA-a-FedAvg-X\% indicates the packet loss ratios (10\%, 30\%, 50\%).}
    \label{fig:evaluation-fairness}
\end{figure*}
\begin{figure*}[!t]
    \centering
    \includegraphics[width=.9\linewidth]{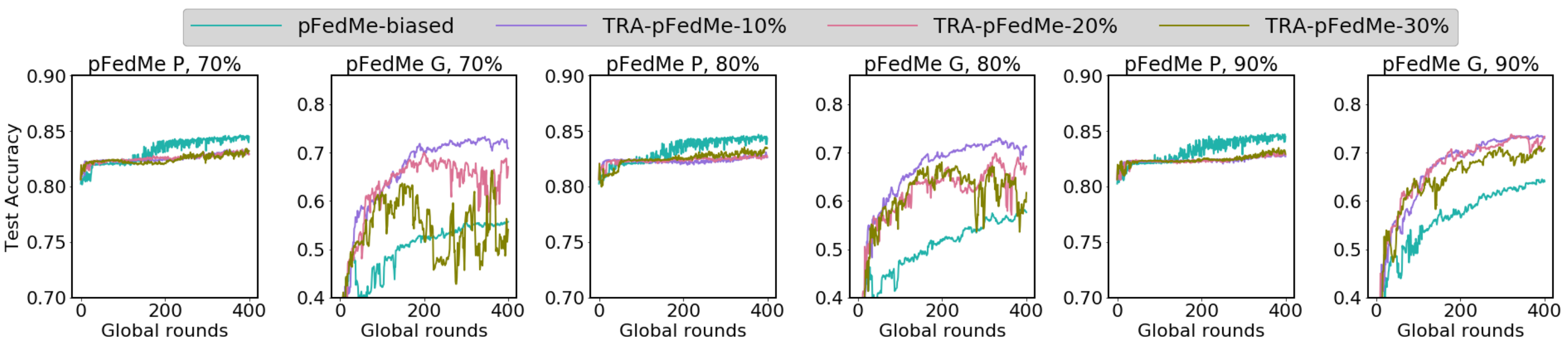}
    \caption{Personalization performance of biased pFedMe and TRA-pFedMe with 70\%, 80\%, and 90\% eligible ratios. TRA-pFedMe-X\% indicates the packet loss ratios (10\%, 20\%, 30\%). We adapted the tested loss ratios according to the observed performance boundary.}
    \label{fig:evaluation-personalization}
\end{figure*}
\paragraph{Aggregation}
Via numerous tests, we find that the combination of \tra and q-FedAvg presents the best aggregation performance in the face of packet loss. As shown in~\cref{fig:evaluation-aggregation}, TRA-q-FedAvg outperforms biased-FedAvg and biased-q-FedAvg in all scenarios. With slightly longer convergence periods, TRA-q-FedAvg (10\% loss ratio) improves the model accuracy on Synthetic(1,1) by 10.35\%/6.69\%, 8.44\%/3.48\%, and 9.31\%/-0.79\%, compared to biased-FedAvg and biased-q-FedAvg in 70\%, 80\%, and 90\% eligible ratio scenarios, respectively. On Synthetic(2,2), the corresponding improvements are 9.88\%/7.39\%, 3.62\%/1.62\%, and 2.75\%/-1.4\%. In a word, when more than 10\% clients have worse network than standard, TRA-q-FedAvg would considerably improve aggregated model accuracy over FedAvg and q-FedAvg using threshold-based selection. We reason that TRA-q-FedAvg presents such improvements thanks to both the features of TRA and q-FedAvg. TRA allows a wider selection of participants thus increasing the learning space with the cost of some data integrity. Meanwhile, q-FedAvg employs the idea of $\alpha$-fairness~\cite{mo2000fair} to give higher relative weights to the clients with higher losses. As such, q-FedAvg compensates for the effect of the packet loss due to TRA. 

\paragraph{Fairness}
We utilize TRA-q-FedAvg to tackle the fairness degradation caused by biased client selection in~\cref{tab:bottleneck-fairness}. 
As shown in~\cref{fig:evaluation-fairness}, TRA-q-FedAvg outperforms biased-q-FedAvg in most scenarios, and the superiority increases as the data heterogeneity increases and the eligible ratio decreases. \cref{tab:performance-fairness} summarizes some numerical results and highlights the best performed algorithms in different scenarios. The results in other scenarios present similar pattern and thus are excluded due to space limit. Note that we use sample based results in~\cref{fig:evaluation-aggregation} to 
measure the accuracy with a higher granularity, while client based results in~\cref{tab:performance-fairness} focus on inter-client fairness. Thus two presented accuracies are different.

\paragraph{Personalization}
We integrate TRA with pFedMe to tackle the personalization performance degradation caused by biased client selection as shown in~\cref{fig:bottleneck-personalization}. As~\cref{fig:evaluation-personalization} depicts, TRA-pFedMe has accuracy of the personal model 1\% lower than pFedMe on average, however outperforms pFedMe in global model accuracy by 20\% at the most. 

\begin{tcolorbox}
\textit{Takeaway:} Integrating TRA with q-FedAvg enables learning from the entire sample space while mitigating the effect of packet loss by adaptively reweighting. As a result, it improves both convergence and fairness performances. TRA considerable improves the global performance of pFedMe compared to in threshold-based settings with a small cost of personalized model accuracy. 
\end{tcolorbox}

\definecolor{Gray}{gray}{0.9}
\begin{table}[!t]
\caption{Client based fairness performance of q-FedAvg with biased selection VS TRA-q-FedAvg with different packet loss ratios. The gray color highlights the best performance algorithms.}
\label{tab:performance-fairness}
\scalebox{0.9}{
\begin{tabular}{lllll}
\specialrule{1.3pt}{1pt}{1pt}
\textbf{Synthetic(1,1)/70\%} & \textbf{Average} & \textbf{Best/Worst 10\%} & \multicolumn{1}{l}{\textbf{Variance}} \\ \hline
\textit{q-FedAvg-biased}  & 55.00\%     & 100\% / 0   & 1439      \\ \hline
\rowcolor{Gray}
\textit{TRA-q-FedAvg-10\%}  & 61.63\%     & 100\% / 6.01\%   & 1031        \\ \hline
\rowcolor{Gray}
\textit{TRA-q-FedAvg-30\%}  & 59.44\%     & 100\% / 4.11\%   & 1021         \\ \hline
\textit{TRA-q-FedAvg-50\%}  & 50.99\%     & 99.97\% / 0   & 1220       \\ \specialrule{1.3pt}{1pt}{1pt}
\textbf{Synthetic(2,2)/70\%} & \textbf{Average} & \textbf{Best/Worst 10\%} & \multicolumn{1}{l}{\textbf{Variance}} \\ \hline
\textit{q-FedAvg-biased}  & 62.34\%     & 100\% / 0   & 1584       \\ \hline
\rowcolor{Gray}
\textit{TRA-q-FedAvg-10\%}  & 69.72\%     & 100\% / 9.81\%   & 870         \\ \hline
\textit{TRA-q-FedAvg-30\%}  & 55.38\%     & 99.69\% / 0   & 1109         \\ \hline
\textit{TRA-q-FedAvg-50\%}  & 55.00\%     & 99.98\% / 2.81\%   & 1125        \\ \hline
\end{tabular}
}
\end{table}


\section{Discussion}
\paragraph{Limitation.}
While TRA-q-FedAvg performs fairly well, we do notice via empirical evaluations that TRA itself does not improve the performance of FedAvg in the face of packet loss. We reason this is because the lightweight recalculation~(Eq.~\eqref{eq:recal}) is not as efficient as q-FedAvg in terms of reweighting. We also note that the performance of TRA can be sensitive to the hyperparameters when combining with some algorithms, e.g., pFedMe. Although we have not uncovered the detailed reasons, we infer it is also due to the limited efficiency of the recalculation.

\paragraph{Future directions.}
Through empirical evaluations, we find that the lightweight TRA works well in lots of scenarios. However we also note that its performance is sensitive to the hyperparameters some times. Besides TRA needs further improvement to guarantee personalized model accuracy while keeping the advantage of global model performance when integrating with personalization algorithms. Therefore next we plan to conduct theoretical analysis of the algorithm and explore its potential with comprehensive optimization problem formulation and solution for bad network tolerance.

\section{Conclusion}
In this work, we investigate loss-tolerant federated learning (LT-FL). Through trace-driven analysis, we find that the commonly assumed limit network challenge is overstated but indeed can cause biased client selection in threshold-based selection settings. We show the bias has severe impacts on different aspects of federated learning. We propose TRA as an complementary solution which allows all clients to participate while toleraing the potential losses of the bad-network clients, who would be filtered out in threshold-based settings. As such TRA balances the model performance and fairness. Integrating TRA with state-of-the-art algorithms shows outperforming performances on aggregation, fairness, and personalization in most scenarios.

\bibliographystyle{named}
\bibliography{ref}

\begin{thebibliography}{}

\bibitem[\protect\citeauthoryear{Barocas \bgroup \em et al.\egroup
  }{2017}]{barocas2017fairness}
Solon Barocas, Moritz Hardt, and Arvind Narayanan.
\newblock Fairness in machine learning.
\newblock {\em NIPS Tutorial}, 1, 2017.

\bibitem[\protect\citeauthoryear{Bonawitz \bgroup \em et al.\egroup
  }{2019}]{bonawitz2019towards}
Keith Bonawitz, Hubert Eichner, Wolfgang Grieskamp, Dzmitry Huba, Alex
  Ingerman, Vladimir Ivanov, Chloe Kiddon, Jakub Kone{\v{c}}n{\`y}, Stefano
  Mazzocchi, H~Brendan McMahan, et~al.
\newblock Towards federated learning at scale: System design.
\newblock {\em arXiv preprint arXiv:1902.01046}, 2019.

\bibitem[\protect\citeauthoryear{Caldas \bgroup \em et al.\egroup
  }{2018}]{caldas2018expanding}
Sebastian Caldas, Jakub Kone{\v{c}}ny, H~Brendan McMahan, and Ameet Talwalkar.
\newblock Expanding the reach of federated learning by reducing client resource
  requirements.
\newblock {\em arXiv preprint arXiv:1812.07210}, 2018.

\bibitem[\protect\citeauthoryear{CCPA}{2021}]{ccpa}
CCPA.
\newblock California consumer privacy act.
\newblock \url{https://www.caprivacy.org/}, 2021.

\bibitem[\protect\citeauthoryear{{Chen} \bgroup \em et al.\egroup
  }{2020}]{9076082}
Y.~{Chen}, X.~{Qin}, J.~{Wang}, C.~{Yu}, and W.~{Gao}.
\newblock Fedhealth: A federated transfer learning framework for wearable
  healthcare.
\newblock {\em IEEE Intelligent Systems}, 2020.

\bibitem[\protect\citeauthoryear{Commission}{2020}]{fccfix}
Federal~Communications Commission.
\newblock {\em Measuring Broadband America Mobile Data}, 2020.
\newblock
  \url{https://www.fcc.gov/reports-research/reports/measuring-broadband-america/measuring-broadband-america-mobile-data}.

\bibitem[\protect\citeauthoryear{Custers \bgroup \em et al.\egroup
  }{2019}]{custers2019eu}
Bart Custers, Alan~M Sears, Francien Dechesne, Ilina Georgieva, Tommaso Tani,
  and Simone Van~der Hof.
\newblock {\em EU Personal Data Protection in Policy and Practice}.
\newblock Springer, 2019.

\bibitem[\protect\citeauthoryear{Dinh \bgroup \em et al.\egroup
  }{2020}]{dinh2020personalized}
Canh~T Dinh, Nguyen~H Tran, and Tuan~Dung Nguyen.
\newblock Personalized federated learning with moreau envelopes.
\newblock {\em Advances in Neural Information Processing Systems 33: Annual
  Conference on Neural Information Processing Systems 2020, NeurIPS 2020},
  2020.

\bibitem[\protect\citeauthoryear{Dong \bgroup \em et al.\egroup
  }{2020}]{dong2020cdc}
Yuanrui Dong, Peng Zhao, Hanqiao Yu, Cong Zhao, and Shusen Yang.
\newblock Cdc: Classification driven compression for bandwidth efficient
  edge-cloud collaborative deep learning.
\newblock {\em Proceedings of the Twenty-Ninth International Joint Conference
  on Artificial Intelligence, IJCAI 20}, pages 3378--3384, 2020.

\bibitem[\protect\citeauthoryear{Fallah \bgroup \em et al.\egroup
  }{2020}]{fallah2020personalized}
Alireza Fallah, Aryan Mokhtari, and Asuman Ozdaglar.
\newblock Personalized federated learning: A meta-learning approach.
\newblock {\em arXiv preprint arXiv:2002.07948}, 2020.

\bibitem[\protect\citeauthoryear{Fang \bgroup \em et al.\egroup
  }{}]{fang20achieving}
Boli Fang, Miao Jiang, Pei-yi Cheng, Jerry Shen, and Yi~Fang.
\newblock Achieving outcome fairness in machine learning models for social
  decision problems.
\newblock In {\em Proceedings of the Twenty-Ninth International Joint
  Conference on Artificial Intelligence, IJCAI-20}, pages 444--450.

\bibitem[\protect\citeauthoryear{Karimireddy \bgroup \em et al.\egroup
  }{2020}]{karimireddy2020scaffold}
Sai~Praneeth Karimireddy, Satyen Kale, Mehryar Mohri, Sashank Reddi, Sebastian
  Stich, and Ananda~Theertha Suresh.
\newblock Scaffold: Stochastic controlled averaging for federated learning.
\newblock In {\em International Conference on Machine Learning}, pages
  5132--5143. PMLR, 2020.

\bibitem[\protect\citeauthoryear{Kone{\v{c}}n{\`y} \bgroup \em et al.\egroup
  }{2016}]{konevcny2016federated}
Jakub Kone{\v{c}}n{\`y}, H~Brendan McMahan, Felix~X Yu, Peter Richt{\'a}rik,
  Ananda~Theertha Suresh, and Dave Bacon.
\newblock Federated learning: Strategies for improving communication
  efficiency.
\newblock {\em arXiv preprint arXiv:1610.05492}, 2016.

\bibitem[\protect\citeauthoryear{Li \bgroup \em et al.\egroup
  }{2019}]{li2019fair}
Tian Li, Maziar Sanjabi, Ahmad Beirami, and Virginia Smith.
\newblock Fair resource allocation in federated learning.
\newblock In {\em International Conference on Learning Representations, ICLR
  2019}, 2019.

\bibitem[\protect\citeauthoryear{Lim \bgroup \em et al.\egroup
  }{2020}]{lim2020federated}
Wei Yang~Bryan Lim, Nguyen~Cong Luong, Dinh~Thai Hoang, Yutao Jiao, Ying-Chang
  Liang, Qiang Yang, Dusit Niyato, and Chunyan Miao.
\newblock Federated learning in mobile edge networks: A comprehensive survey.
\newblock {\em IEEE Communications Surveys \& Tutorials}, 2020.

\bibitem[\protect\citeauthoryear{Lin \bgroup \em et al.\egroup
  }{2020}]{lin2020federated}
Frank Po-Chen Lin, Christopher~G Brinton, and Nicolo Michelusi.
\newblock Federated learning with communication delay in edge networks.
\newblock {\em arXiv preprint arXiv:2008.09323}, 2020.

\bibitem[\protect\citeauthoryear{Lyu \bgroup \em et al.\egroup
  }{2020}]{lyu2020collaborative}
Lingjuan Lyu, Xinyi Xu, Qian Wang, and Han Yu.
\newblock Collaborative fairness in federated learning.
\newblock In {\em Federated Learning}, pages 189--204. Springer, 2020.

\bibitem[\protect\citeauthoryear{McMahan \bgroup \em et al.\egroup
  }{2017}]{mcmahan2017communication}
Brendan McMahan, Eider Moore, Daniel Ramage, Seth Hampson, and Blaise~Aguera
  Arcas.
\newblock Communication-efficient learning of deep networks from decentralized
  data.
\newblock In {\em Artificial Intelligence and Statistics}. PMLR, 2017.

\bibitem[\protect\citeauthoryear{Mo and Walrand}{2000}]{mo2000fair}
Jeonghoon Mo and Jean Walrand.
\newblock Fair end-to-end window-based congestion control.
\newblock {\em IEEE/ACM Transactions on networking}, 2000.

\bibitem[\protect\citeauthoryear{Mohri \bgroup \em et al.\egroup
  }{2019}]{mohri2019agnostic}
Mehryar Mohri, Gary Sivek, and Ananda~Theertha Suresh.
\newblock Agnostic federated learning.
\newblock In {\em International Conference on Machine Learning, ICML 2019},
  pages 4615--4625, 2019.

\bibitem[\protect\citeauthoryear{Nishio and Yonetani}{2019}]{nishio2019client}
Takayuki Nishio and Ryo Yonetani.
\newblock Client selection for federated learning with heterogeneous resources
  in mobile edge.
\newblock In {\em ICC}. IEEE, 2019.

\bibitem[\protect\citeauthoryear{Openmined}{2021}]{openmined}
Openmined.
\newblock Openmined.
\newblock \url{https://www.openmined.org/}, 2021.

\bibitem[\protect\citeauthoryear{Xia \bgroup \em et al.\egroup
  }{2019}]{xia2019rethinking}
Jiacheng Xia, Gaoxiong Zeng, Junxue Zhang, Weiyan Wang, Wei Bai, Junchen Jiang,
  and Kai Chen.
\newblock Rethinking transport layer design for distributed machine learning.
\newblock In {\em APNet}, 2019.

\end{thebibliography}

\end{document}